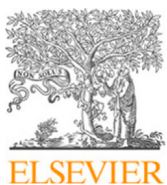
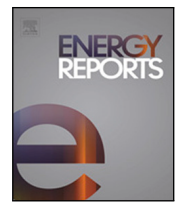

Research paper

# Kolmogorov–Arnold recurrent network for short term load forecasting across diverse consumers

Muhammad Umair Danish, Katarina Grolinger *

*Electrical and Computer Engineering, Western University, 1151 Richmond St, London, N6A 3K7, Ontario, Canada*

A B S T R A C T

Load forecasting plays a crucial role in energy management, directly impacting grid stability, operational efficiency, cost reduction, and environmental sustainability. Traditional Vanilla Recurrent Neural Networks (RNNs) face issues such as vanishing and exploding gradients, whereas sophisticated RNNs such as Long Short-Term Memory Networks (LSTMs) have shown considerable success in this domain. However, these models often struggle to accurately capture complex and sudden variations in energy consumption, and their applicability is typically limited to specific consumer types, such as offices or schools. To address these challenges, this paper proposes the Kolmogorov–Arnold Recurrent Network (KARN), a novel load forecasting approach that combines the flexibility of Kolmogorov–Arnold Networks with RNN's temporal modeling capabilities. KARN utilizes learnable temporal spline functions and edge-based activations to better model non-linear relationships in load data, making it adaptable across a diverse range of consumer types. The proposed KARN model was rigorously evaluated on a variety of real-world datasets, including student residences, detached homes, a home with electric vehicle charging, a townhouse, and industrial buildings. Across all these consumer categories, KARN consistently outperformed traditional Vanilla RNNs, while it surpassed LSTM and Gated Recurrent Units (GRUs) in six buildings. The results demonstrate KARN's superior accuracy and applicability, making it a promising tool for enhancing load forecasting in diverse energy management scenarios.

## 1. Introduction

The global electricity demand is projected to grow by 4% in 2025, up from 2.5% in 2023, with renewable sources expected to contribute 35% of the total share (Business Standard News, 2024). This would represent the highest annual growth since 2007. Accurate energy load forecasting not only aids in energy management and delivery but also contributes to financial savings and achieving climate goals (Business Standard News, 2024). For example, improving energy efficiency alone could save U.S. consumers $450 billion in electric bills by 2030 (Business Standard News, 2024). Moreover, accurate load forecasting is instrumental in reaching climate goals, such as the European Green Deal's target to reduce carbon emissions by 55% and improve energy efficiency by 32.5% by 2030 (European Climate, Energy and Environment, 2024). This emphasizes the indispensable role of accurate load forecasting for both economic and environmental sustainability.

Given the important role of load forecasting in ensuring the stability and reliability of energy supply, short-term load forecasting has evolved as a pivotal area of focus in both research and practical applications (Deng et al., 2021). Short term load forecasting enables utilities to promptly respond to fluctuations in demand, which is vital for maintaining grid stability and operational efficiency. Numerous artificial intelligence (AI) and Machine Learning (ML) technologies (Muzaffar and Afshari, 2019; Yao et al., 2022; Skala et al., 2023; Qin et al., 2022) have been used to improve load forecasting accuracy. However, despite the advancements brought by these technologies, accurately forecasting load in diverse urban settings remains a challenge due to factors such as consumer usage behavior and meteorological shifts (Chen et al., 2021). Among various AI technologies, Deep Learning (DL) models, including Recurrent Neural Networks (RNNs), Long Short-Term Memory Networks (LSTMs), Gated Recurrent Units (GRUs), and Transformers, have been recognized for their ability to capture long-range temporal dependencies (L'Heureux et al., 2022).

Vanilla RNNs were the earliest neural networks designed to handle sequential data, but they have challenges such as vanishing and exploding gradients and difficulties in learning long-term dependencies. These

---

☆ This work was supported in part by the Climate Action and Awareness Fund [EDF-CA-2021i018, Environment Canada, K. Grolinger] and in part by the Canada Research Chairs Program [CRC-2022-00078, K. Grolinger].
 * Corresponding author.
   *E-mail address:* kgroling@uwo.ca (K. Grolinger).

https://doi.org/10.1016/j.egyr.2024.12.038
Received 28 September 2024; Received in revised form 5 December 2024; Accepted 13 December 2024
Available online 24 December 2024

challenges led to the development of more sophisticated recurrent architectures, such as LSTM networks and GRUs (Yu et al., 2019). These recurrent DL models share foundational principles with Multi-Layer Perceptrons (MLPs), such as the use of fully connected linear weight matrices and fixed non-linear activation functions (such as Sigmoid or Tanh) at the nodes or neurons, enabling them to approximate complex functions (Li et al., 2021). This structure imposes a fixed non-linearity by using predetermined and unchanging activation functions that provide a constant level of non-linearity, which is not always sufficient to capture the complex relationships in load patterns for diverse consumer types, given their variable nature of energy usage influenced by sudden events, lifestyle changes, and meteorological shifts (Liu et al., 2024; Cai et al., 2019). Moreover, these models have been largely confined to specific consumer categories, such as residences (Gong et al., 2021), apartments (Rezaei and Dagdougui, 2020), detached houses (Kong et al., 2017), townhouses (Gong et al., 2021), and homes with electric vehicles (Zhang et al., 2020). However, a comprehensive assessment across diverse consumer settings is necessary, as consumer behavior varies by building type. This highlights the need for studies that extend analyses to a broader range of consumer types to ensure the robustness of forecasting models.

Recently, a new AI technology, the Kolmogorov–Arnold Network (KAN), has been introduced as an alternative to MLPs (Liu et al., 2024). Unlike MLPs, KANs have learnable activations on edges or weights instead of predefined activations on nodes or neurons. This means that activations in KANs are not applied over the weights and biases but are instead applied separately in a learnable manner on the edges. KANs do not have linear weights at all; instead, every weight matrix is replaced by a univariate function parameterized as a spline, making them less likely to face vanishing and exploding gradients (Liu et al., 2024). These univariate functions are learnable and provide a more flexible and potentially more accurate way of representing relationships in the data (Liu et al., 2024). The current design of KAN (Liu et al., 2024) is not well-suited for handling time series load data as they were designed for tabular data without considering the time component. This assumption makes KANs less suitable for temporal data because they lacks the ability to capture long-term dependencies, which is essential for accurate time series load forecasting.

To take advantage of KANs for load forecasting, this paper proposes Kolmogorov–Arnold Recurrent Network (KARN), a temporal version of KAN that combines the flexibility of learnable activations on edges with the ability of RNNs to capture long-term dependencies. This allows for greater adaptability to varying consumer behaviors and external influences, without the risk of vanishing or exploding gradients. This paper makes the following main contributions:

1. Design of KARN, which is based on the Kolmogorov–Arnold Representation Theorem and incorporates a recurrent structure to retain the memory of previous states, with an emphasis on improving the accuracy of load forecasting across diverse consumer types.
2. Design of the temporal basis and temporal spline functions to enhance KARN's ability to capture long-range dependencies.
3. Comprehensive analysis of KARN across diverse consumer types, including residences, detached houses, townhouses, houses with electric vehicles, commercial and industrial datasets, and comparison with traditional recurrent neural networks.

The remainder of this paper is organized as follows: Section 2 presents the related work, Section 3 details the architecture of KARN and the mathematical foundations underlying their design. Section 4 describes the experimental setup, including the datasets used and the evaluation metrics. Section 5 discusses the results and compares the performance of KARN with other state-of-the-art models. Finally, Section 6 concludes the paper and suggests directions for future research.

## 2. Related work

This section provides an overview of the notable advancements in load forecasting followed by a discussion on applications of KANs.

### 2.1. Load forecasting with deep learning

To enhance load forecasting accuracy, various DL approaches have been proposed and extensively studied. The most notable models are Vanilla RNNs, LSTM networks, GRUs, and Transformers (Zeng et al., 2023). Vanilla RNN was the earliest model designed for handling sequential and time series data, but it faces several challenges, including vanishing gradients, exploding gradients, gradient instability during backpropagation through time, difficulty in retaining memory of long past time steps and overfitting, making it difficult to capture complex, long-range dependencies (Yu et al., 2019). These issues in Vanilla RNNs led to the development of LSTM and GRU, which utilize multiple recurrent gates to capture memory over longer time steps. LSTM and GRU, along with their hybrid and enhanced variants, are currently at the forefront of research in the field of load forecasting (Skala et al., 2023; Bessani et al., 2020; Wu and Peng, 2017; Zhang and Chiang, 2019; Yamak et al., 2019). For instance, Skala et al. (2023) introduced LSTM Bayesian neural networks for interval load forecasting in the context of individual households with electric vehicle charging by demonstrating significant improvements in forecasting precision. Bessani et al. (2020) proposed Bayesian neural networks for individual houses and demonstrated that the proposed method outperforms traditional neural networks.

Wu and Peng (2017) applied k-means clustering and bagging techniques with neural networks to enhance the accuracy of short-term forecasting for wind turbines. He et al. (2024) proposed a hybrid model consisting of Holt–Winters and GRU for short-term load-interval forecasting using the 2022 Teddy Cup dataset and demonstrated performance improvements over traditional models. Ahmed and Jamil (2024) deployed vanilla RNN, LSTM, GRU, and their bi-directional variants on a university campus building electricity load dataset, finding that the bi-directional GRU outperformed others for this specific dataset. Wei et al. (2024) proposed a transfer domain selection algorithm that integrates the Wasserstein distance and maximal information coefficient and validated it on a public office dataset based in Greece. Among RNNs, LSTM has been particularly successful in load forecasting due to its gating mechanisms, which allow it to capture long-term dependencies (Yamak et al., 2019). This capability is crucial in short-term load forecasting, where sudden fluctuations in power demand are common.

Transformers and their advanced variants have enhanced prediction capabilities through self-attention mechanisms that effectively handle complex data patterns (Vaswani et al., 2017). L'Heureux et al. (2022) proposed a transformer-based architecture and evaluated it on an open-source dataset containing data from 20 zones of a U.S. utility company. The findings showed that the transformer outperformed both LSTM and sequence-to-sequence models in terms of predictive accuracy. Mao et al. (2024) proposed a hybrid model combining GRU and Transformer architectures, specifically designed for a dataset from Panama City. The study showed that this hybrid approach delivered performance improvements on certain datasets. Giacomazzi et al. (2023) employed the Temporal Fusion Transformer (TFT) for load forecasting using the Hanoi city datasets and found that the newly developed TFT outperformed both LSTMs and traditional Transformers in this context. Kong et al. (2017) proposed the Time Augmented Transformer (TAT), a hybrid variant of the transformer model. In TAT, a time augmentation module is introduced, which uses one-hot encoding followed by fully connected layers. This model was applied to real-world residential buildings and demonstrated superior performance compared to seq2seq and LSTM models.

The evolution of hybrid models, which integrate different approaches, has become increasingly prominent in the field of load forecasting (Li et al., 2023a,b; Triebe et al., 2019; Danish and Grolinger, 2024). Tan et al. (2024) proposed an innovative hybrid model called InE-BiLSTM, which combines an Informer Encoder with bidirectional LSTM. This model was deployed for two public building datasets and demonstrated significant improvements in performance, albeit with an increase in computational cost. E-ELITE neural network proposed by Zhang and Chiang (2019) for short-term load forecasting for industrial applications using the ISO New England dataset, achieved superior performance compared to traditional models.

The aforementioned DL models are grounded in the Universal Approximation Theorem, which involves linear weights combined with non-linear activations applied to neurons. While effective for many tasks, this design can make their applicability challenging for certain load datasets, particularly those with high variability and unpredictability due to consumer usage patterns. Additionally, these aforementioned DL-based studies have achieved significant success in load forecasting, but their accuracy has not been extensively assessed or validated across various consumer types. This limitation arises from the distinct electricity usage patterns among different consumers, which can introduce spikes, drops, concept drifts, and level shifts in the data. Such variability can reduce forecasting performance, especially in datasets where randomness is driven by consumer behavior. This gap raises an important question: Can a single forecasting model effectively adapt to and capture the diverse energy consumption patterns demonstrated across various consumer groups?

To address these challenges, our study proposes KARN which is designed to model complex patterns by leveraging the flexibility of KANs with the temporal modeling capabilities of recurrent structures. KARN overcomes the limitations of traditional DL models by utilizing learnable activation functions on edges, rather than nodes, allowing for a more adaptable and precise representation of relationships in the data. Unlike traditional RNNs, KARN's structure inherently mitigates issues such as vanishing and exploding gradients, as the spline-based activations on edges offer smoother gradient flows and more robust learning dynamics. This design is particularly well-suited to capturing the diverse and dynamic energy consumption patterns across various consumer types, leading to superior performance compared to traditional RNNs.

## 2.2. Kolmogorov–Arnold networks

KANs have emerged as a promising alternative to traditional MLP (Liu et al., 2024). Unlike MLPs, which are based on the Universal Approximation Theorem, KANs are inspired by the Kolmogorov–Arnold representation theorem. This key difference allows KANs to replace linear weights with univariate functions parameterized as splines, enabling the placement of learnable activations on edges or weights rather than on nodes or neurons. The ability of KANs to adapt to complex data patterns without relying on fixed activation functions has positioned them as a valuable tool in scientific computing and data analysis (Vaca-Rubio et al., 2024; Genet and Inzirillo, 2024b,a). This design provides KANs with a greater number of learnable parameters, which can lead to improved performance at the cost of increased computational complexity.

Since Liu et al. (2024) introduced KANs, their application has been increasingly explored across various fields, demonstrating advantages over conventional ML technologies. For example, Vaca-Rubio et al. (2024) investigated the use of KANs for time series traffic forecasting and compared their performance with that of MLPs across different layer configurations. The study found KANs to be more powerful predictors; however, this implementation used standard KANs and MLPs without modifying KANs to accept temporal inputs, which can negatively impact the ability of KANs to capture the temporal nature of the data, potentially limiting their forecasting accuracy in time-dependent tasks.

Genet and Inzirillo (2024b) introduced Temporal KAN (TKAN) for time series forecasting on Binance cryptocurrency data, aiming to develop a model comparable to LSTM. The adaptation of KANs for temporal data represents a step forward; however, this strategy incorporated linear weights and fully connected layers, which deviates somewhat from the original KAN architecture that emphasizes non-linear spline-based weight matrices. This integration dilutes some of the distinctive advantages that KANs offer through their non-linear spline-based weights. It may be more appropriate to consider this model as a hybrid rather than a direct equivalent of an LSTM based purely on KAN principles, as it introduces elements beyond the traditional KAN framework. Similarly, Genet and Inzirillo (2024a) also proposed the Temporal Kolmogorov–Arnold Transformer (TKAT), which is analogous to the Temporal Fusion Transformer but uses TKANs in its implementation. This can also be seen as a hybrid model since it once again incorporates linear weights from MLPs, thereby not fully exploring the potential accuracy offered by non-linear spline-based weights alone. Xu et al. (2024) proposed MT-KAN for financial time series datasets, focusing specifically on interpretability by incorporating symbolic regression. Their findings demonstrated that KANs were more interpretable compared to traditional AI technologies and also outperformed MLPs in terms of accuracy (Zhang et al., 2022).

KANs have also been explored across various domains beyond time series data. Azam and Akhtar (2024) evaluated the applicability of KANs to computer vision tasks, identifying challenges in their effectiveness for image recognition. Cheon (2024) utilized KANs for satellite image classification and reported slightly better performance compared to traditional MLPs. Kiamari et al. (2024) introduced GKAN (Graph KAN) as a substitute for Graph Convolutional Neural Networks, achieving improved accuracy at the cost of increased computational requirements. Tang et al. (2024) developed U-KAN, a variant analogous to the traditional U-Net, for brain image segmentation, and observed marginally better performance compared to U-Net. Aghaei (2024) proposed Rational KAN (rKAN) and evaluated the performance of rKAN against various DL and physics-informed deep learning models. The results showed that rKAN outperformed these models.

Despite the diverse applications of KANs across various domains, their use in load forecasting remains unexplored. While the existing literature has demonstrated the effectiveness of KANs in tasks such as time series forecasting, computer vision, and graph-based modeling, the KANs potential in load forecasting for diverse consumer types has not been investigated. This gap underscores the need to explore how KANs can be adapted and applied to the specific challenges of energy load forecasting.

To address the limitations, our study designs a recurrent variant of KAN, aimed at enhancing load forecasting accuracy. Unlike traditional RNNs, which are prone to vanishing and exploding gradients, this study proposes KARN which leverages the inherent flexibility of basis splines and learnable independent activations on edges. This learning mechanism minimizes the risk of gradient issues, as activations are not applied directly to weights, biases, or spline coefficients. By integrating the strengths of KANs with a recurrent structure, KARN effectively captures temporal dependencies in load data while ensuring more robust learning dynamics. This innovative approach positions KARN as a pioneering solution in the field of load forecasting for diverse consumer types.

## 3. Kolmogorov–Arnold recurrent network

This section presents the Kolmogorov–Arnold Recurrent Network (KARN), our proposed method for load forecasting as illustrated in Fig. 1, designed to accommodate a wide range of energy consumers. KARN extends KAN to handle time series energy data. It is based on the Kolmogorov–Arnold representation theorem, which offers a method to represent multivariate continuous functions using a finite composition of univariate functions, which is achieved by placing learnable activation functions on weights, parameterized as splines. The KARN is designed to process the current input $x_{(t)}$ at time step $t$ and keep the memory of the previous time steps. This allows KARN to calculate the next hidden state and make forecasts. Several learnable weight parameters additionally aid the temporal design. This temporal design allows KARN to approximate functions in a recurrent manner while retaining the advantages of KANs, such as non-linear weight metrics and activations on weights through the use of basis and spline functions. These features render KARN highly suitable for load forecasting across a diverse set of consumers.

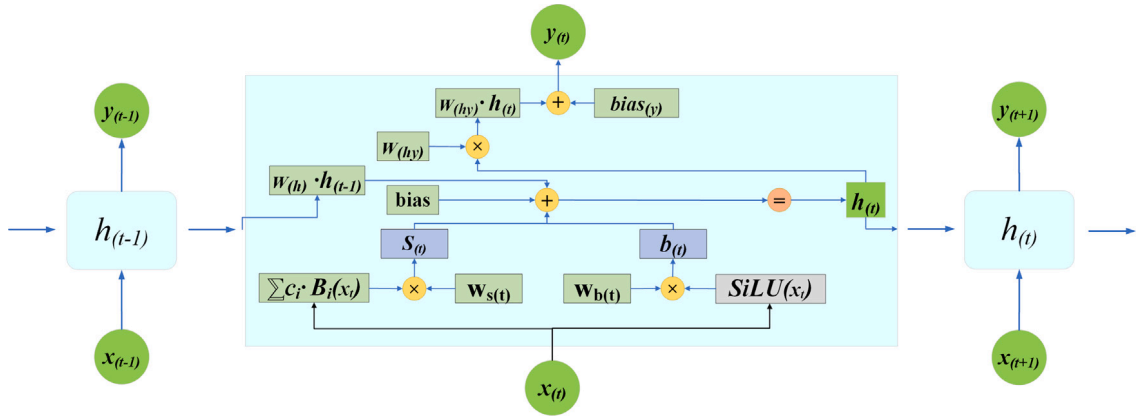

**Fig. 1.** The proposed KARN architecture, a deep learning recurrent model, consists of three main components: (a) the temporal basis function, represented by the SiLU transformation applied to the input and weighted through $W_{b(t)}$, producing $b_{(t)}$; (b) the temporal spline function, applied to a B-spline basis, weighted through $W_{s(t)}$, producing $S_{(t)}$; and (c) the hidden state update mechanism, where the outputs of the basis and spline functions, along with a learnable bias, and previous hidden state are combined to form the updated hidden state $h_{(t)}$. The hidden state is then used to compute the output $y_{(t)}$, ensuring sequential dependencies are maintained throughout the time series.

In our case of load forecasting, the input tensor is $x \in \mathbb{R}^{m \times f \times n}$, where $m$ represents the number of samples, $f$ denotes the number of features, and $n$ indicates the number of time steps within each sample. These time steps are generated using the sliding window technique (L'Heureux et al., 2022), which creates sequence segments from the time series energy data. The features $f$ typically include variables such as temperature, day of the year, day of the month, day of the week, hour of the day, and energy consumption from the preceding $n$ hours. The following subsections will describe the temporal basis function, temporal spline function, memory mechanism in KARN, grid extension, and loss function.

### 3.1. Temporal basis function

The temporal basis function introduces non-linearity into the model and is chosen to provide a simple, smooth, and well-behaved nonlinear transformation of the input. In the standard KAN variant (Liu et al., 2024), the *SiLU* (Sigmoid Linear Unit) function is employed. The basis function can be viewed as a residual connection that helps stabilize the learning process. The SiLU function is defined as:

$$\text{SiLU}(x) = \frac{x}{1 + e^{-x}} \tag{1}$$

This function is used to improve gradient flow and learning efficiency in temporal load data. The temporal basis function takes input data $\mathbf{x} \in \mathbb{R}^{m \times f \times n}$ over time $t$, applies the SiLU function to transform each time step, enabling the model to extract meaningful patterns from the temporal structure of the input. The transformation at each time step $t$ is given by:

$$b_{(t)} = \mathbf{w}_{b(t)} \cdot \text{SiLU}(\mathbf{x}_{(t)}) \tag{2}$$

where $\mathbf{w}_{b(t)}$ are learnable weights corresponding to the input at time $t$ that allow the model to dynamically adapt to temporal data patterns. The temporal basis function ensures that the model can capture non-linear transformations that evolve over time by enabling it to process time-varying inputs.

### 3.2. Temporal spline function

The temporal spline function is a piecewise polynomial function that models complex, non-linear relationships by transforming input data using a series of B-spline functions. Each spline or piecewise polynomial function is defined over a specific grid interval. A grid interval, also known as a knot, refers to the set of points, where the spline functions are evaluated and applied. These grid points divide the input space into segments where spline functions change or adapt, allowing the model to capture the non-linear relationships within data. The positioning and number of these grid points can affect the accuracy of the spline function approximation as they determine where and how the splines will adjust to the underlying data patterns (De Boor, 1972; Liu et al., 2024). The degree of polynomial in a spline is the highest power of the variable in each polynomial segment between the grid points.

The temporal spline function extends the spline to operate in recurrent manners. These basis splines take input as $\mathbf{x}_{(t)}$ over time $t$ and apply a piecewise polynomial transformation of a degree $p$, defined over grid points $G$ that divide inputs space. The transformation is controlled by learnable coefficients $\mathbf{c}_i$ to capture underlying patterns in the data. The learnable coefficients allow for the dynamic adjustment of the influence of each spline function at time step $t$, enabling the model to capture long-term dependencies in the data. The output of these spline functions is then aggregated to form a complete representation. The transformation at each time step $t$ is given by:

$$S_{(t)} = \mathbf{w}_{s(t)} \cdot \sum_i \mathbf{c}_i \cdot B_i(\mathbf{x}_{(t)}) \tag{3}$$

In this expression, $B_i$ represents the $i$th B-spline basis function, $\mathbf{c}_i$ are the learnable coefficients that adjust the contribution of each spline, and $\mathbf{w}_{s(t)}$ are the learnable weights applied to the aggregated output of the spline functions. This means that the temporal spline function in KARN has two sets of learnable parameters: weights that control the effects after aggregation and coefficients that learn patterns before aggregation to provide a fine-tuned learning mechanism. The temporal spline function ensures that the model can capture evolving, non-linear relationships in the load data by dynamically adapting the splines to changing input patterns over time.

### 3.3. Memory mechanism in KARN

The previous two subsections explained how the basic components are extended to handle temporal data. This subsection explains how to combine both components to formulate the KARN layer. The KARN layer is the combination of both the temporal basis function and the temporal spline function, along with the contribution from the previous hidden state. The hidden state at each time step $h_{(t)}$ is calculated by combining the outputs of the temporal basis function, the temporal spline function, and the hidden state from the previous time step, along with learnable weight metrics and a bias term:

$$h_{(t)} = W_{(h)} \cdot h_{(t-1)} + b_{(t)} + S_{(t)} + \text{bias} \tag{4}$$

In the above equation, $W_{(h)}$ is a learnable weight matrix that helps retain information from the previous hidden state $h_{(t-1)}$ to the current state $h_{(t)}$. The term $b_{(t)}$ represents the output of the temporal basis function applied to the input at time step $t$, while $S_{(t)}$ is the output of the temporal spline function applied to the same input. The bias term is a learnable parameter that adjusts the final hidden state. This recurrent mechanism allows the model to keep memory from previous time steps while processing the current input, enabling KARN to capture both short-term dependencies (through $x_{(t)}$) and long-term dependencies (through $h_{(t-1)}, h_{(t-2)}, \ldots$). At each time step $t$, the hidden state $h_{(t)}$ is updated based on both the current input $x_{(t)}$ and the hidden state from the previous time step $h_{(t-1)}$, allowing the model to propagate information forward through time and dynamically adapt to new inputs, as illustrated in Fig. 1.

The model generates an output at each time step based on the hidden state at that time. The output $y_{(t)}$ is computed by applying learnable weights to the hidden state $h_{(t)}$:

$$y_{(t)} = \mathbf{W}_{(hy)} \cdot h_t + \text{bias}_{(y)} \tag{5}$$

Here $\mathbf{W}_{(hy)}$ is a learnable weight matrix that maps the hidden state to the output space while the bias term $bias_{(y)}$ adjusts the output. This ensures that the model can generate an output at each time step, leveraging the information stored in the hidden state, which has been influenced by both the current and previous inputs. Similar to the standard KAN (Liu et al., 2024), weight sharing is also utilized, which allows multiple spline functions to use the same set of parameters, reducing the total number of learnable parameters and enhancing the model's abilities. Additionally, a locking mechanism is in place that manages and enforces parameter sharing across spline functions, ensuring that once parameters are shared, they remain consistently applied throughout the training process.

### 3.4. Grid extension

The B-spline $B_i(x)$ of degree $p$ is defined recursively over intervals known as knots. The first step in applying B-splines is to define a sequence of grid points which divide the domain into intervals. The grid is initialized based on a specified range (e.g., $[-1, 1]$) and is divided into intervals according to the number of grid points. Within each interval, the B-spline acts as a piecewise polynomial of a degree $p$, making B-splines well-suited for use in neural networks. For a given set of knots $\{\tau_0, \tau_1, \ldots, \tau_n\}$, the B-spline of degree $p$ is defined as:

$$B_i^p(x) = \frac{x - \tau_i}{\tau_{i+p} - \tau_i} B_i^{p-1}(x) + \frac{\tau_{i+p+1} - x}{\tau_{i+p+1} - \tau_{i+1}} B_{i+1}^{p-1}(x) \tag{6}$$

In this equation, $B_i^p(x)$ equals 1 if $\tau_i \le x < \tau_{i+1}$, and 0 otherwise. The first term $\frac{x-\tau_i}{\tau_{i+p}-\tau_i} B_i^{p-1}(x)$ handles the contribution of the $i$th basis function over its interval, while the second term addresses the contribution of the previous basis function.

Grid extension is a technique that adds more points to the grid or knots. Since splines are defined on these grid points, increasing the number of points will result in more splines and each spline has an associated learnable coefficient $c_i$. This means that as the grid is extended, the number of learnable coefficients $c_i$ and splines increases, ultimately enhancing the model's ability to learn complex data patterns. The process of grid extension allows KARN trained with fewer parameters (a coarser grid) to be expanded to a model with more parameters (a finer grid) without the need to retrain the model from scratch. Consider the task of approximating a function $f(x)$ over a bounded interval $[a, b]$ using B-splines of order $p$. Initially, the approximation is constructed on a coarse grid with $G_1$ intervals, defined by grid points $\{\tau_0 = a, \tau_1, \ldots, \tau_{G_1} = b\}$. The function is then expressed as a linear combination of the corresponding B-spline basis functions:

$$f_{coarse}(x) = \sum_{i=0}^{G_1+p-1} c_i \cdot B_i(x) \tag{7}$$

To enhance the accuracy of the approximation, a finer grid with $G_2$ intervals is used. The function is subsequently re-expressed on this refined grid as:

$$f_{fine}(x) = \sum_{j=0}^{G_2+p-1} c'_j \cdot B'_j(x) \tag{8}$$

The new coefficients $c'_j$ associated with the fine grid are obtained by minimizing the discrepancy between $f_{fine}(x)$ and $f_{coarse}(x)$ over the distribution of $x$:

$$\{c'_j\} = \arg\min_{\{c'_j\}} \mathbb{E}_{x \sim p(x)} \left[ \sum_{j=0}^{G_2+p-1} c'_j \cdot B'_j(x) - \sum_{i=0}^{G_1+p-1} c_i \cdot B_i(x) \right]^2 \tag{9}$$

This optimization can be effectively solved using a least squares algorithm, allowing the grid to be extended without necessitating the retraining of the model (Liu et al., 2024).

### 3.5. Loss function

The Mean Squared Error (MSE) or Mean Absolute Error (MAE) loss functions, determined through hyperparameter optimization, are used to quantify the difference between the predicted and actual target values. These are calculated as:

$$\mathcal{L}_{MSE} = \frac{1}{m} \sum_{i=1}^{m} (\hat{y}_i - y_i)^2, \tag{10}$$

$$\mathcal{L}_{MAE} = \frac{1}{m} \sum_{i=1}^{m} |\hat{y}_i - y_i|, \tag{11}$$

where $\hat{y}_i$ represents the predicted output, $y_i$ denotes the actual target value, and $m$ is the number of samples. It is important to note that MAE is more robust to outliers than MSE because MSE squares the errors, amplifying the impact of outliers. However, MSE provides a smoother gradient for optimization, facilitating more controlled updates during training. Consequently, the choice between the two is made during the hyperparameter optimization process.

## 4. Evaluation

This section describes the datasets, preprocessing steps, performance metrics, hyperparameter optimization, and the architectures included in the comparison.

### 4.1. Datasets and preprocessing

The evaluation employed ten distinct real-world datasets from three primary consumer groups: student residences, individual homes, and industrial and commercial buildings. Table 1 provides an overview of these datasets, including the time frames during which data were collected and a brief description of each. There are significant differences between the building's energy consumption. For example, Residence 1 has suite-style accommodation with a shared kitchen, while Residence 2 follows a suite-style but lacks a kitchen. Both residences accommodate over 400 students.

Although all the homes are located in London, Ontario, Canada, there is considerable diversity among them. Homes 1, 2, and 3 are detached properties, but Home 3 is distinct due to the presence of an electric vehicle charging, resulting in notable load fluctuations caused by at-home charging. Home 4 is a 3-bedroom townhouse, where energy

**Table 1**
Energy consumption datasets used in evaluation.

| Dataset | Dates | Short description |
|---|---|---|
| **Student residences** | | |
| Residence 1 | Jan/2019–Jul/2023 | A suite-style residence with shared kitchen |
| Residence 2 | Jan/2019–Jul/2023 | A suite-style residence without a kitchen |
| **Individual houses** | | |
| House 1 | Jan/2002–Dec/2004 | A detached home with complex energy usage patterns |
| House 2 | Mar/2021–Aug/2021 | A 2-bedroom detached house |
| House 3 | Mar/2021–Aug/2021 | A 2-bedroom detached house with an electric vehicle |
| House 4 | Mar/2021–Aug/2021 | A 3-bedroom townhouse |
| **Industrial and commercial** | | |
| Manufacturing | Jan/2016–Dec/2017 | A manufacturing unit |
| Medical clinic | Jan/2016–Dec/2017 | A medical and wellness clinic |
| Retail store | Jan/2016–Dec/2017 | A retail store |
| Office | Jan/2016–Dec/2017 | A dedicated office building |

consumption patterns differ from detached homes due to the influence of neighboring units.

Each dataset has a date/time and hourly energy consumption. From the date/time, additional features were extracted including the day of the year, the day of the month, the day of the week, and the hour of the day to assist in modeling seasonal, weekly, and daily patterns. To capture weather patterns, temperature data was added and additional relevant features can be integrated if available. To minimize the impact of large-scale features and enhance convergence, Min-Max scaling was applied.

The datasets were split into training, validation, and test sets in a 60%-20%–20% ratio. Temporal data for the models was prepared using a sliding window technique with a window length of 24 and a stride of 1. All models were provided with the previous 24 h of five features, including energy load, and were tasked with predicting the next 24 h of energy load. This forecasting length was selected because energy operations commonly rely on next-day forecasts for energy planning.

### 4.2. Evaluation metrics

The evaluation was conducted using three metrics: Mean Absolute Error (MAE), Root Mean Square Error (RMSE), and Symmetric Mean Absolute Percentage Error (SMAPE). These metrics were selected to provide a comprehensive assessment of the model's performance from different perspectives.

MAE measures the average absolute difference between predicted and actual values, which is a straightforward interpretation of the error in forecasting. MAE is useful for understanding the general accuracy of the model across the entire dataset (Fekri et al., 2023). MAE is calculated as:

$$\text{MAE} = \frac{1}{m} \sum_{i=1}^{m} |y_i - \hat{y}_i|, \quad (12)$$

where $y_i$ represents the actual load values, $\hat{y}_i$ represents the predicted load values, and $m$ is the number of samples.

RMSE is more sensitive to large errors than MAE due to the squaring of the residuals before averaging, making it a valuable metric when large errors are undesirable (Fekri et al., 2023; L'Heureux et al., 2022). RMSE is calculated as:

$$\text{RMSE} = \sqrt{\frac{1}{m} \sum_{i=1}^{m} (y_i - \hat{y}_i)^2}, \quad (13)$$

SMAPE expresses the forecasting error as a percentage, facilitating easy interpretation and enabling comparison across different datasets. SMAPE was selected over Mean Absolute Percentage Error (MAPE) because MAPE is biased toward large values and becomes undefined when actual values are zero. SMAPE addresses these issues by symmetrizing the error calculation, providing a more robust and interpretable metric

**Table 2**
Forecasting techniques included in comparison.

| No. | Model | Architecture |
|---|---|---|
| 1 | RNN | Vanilla RNN with fully connected layer. |
| 2 | GRU | Standard GRU with fully connected layer. |
| 3 | LSTM | Standard LSTM with fully connected layer. |

**Table 3**
Hyperparameter search space for all models.

| Hyperparameter | Range of values |
|---|---|
| Size of hidden Layer (all models) | 64, 128, 256 |
| No. of layer (all models) | 1, 2, 3 |
| Choice of optimizer (all models) | Adam, SGD, AdamW |
| Objective function (all models) | MAE, MSE |
| Degree $k$ (our model only) | 1, 2, 3 |
| Grid points (our model only) | [2–14] |

in cases where the data include zero or near-zero values. SMAPE metric is calculated as:

$$\text{SMAPE} = 100\% \times \frac{1}{m} \sum_{i=1}^{m} \frac{2|y_i - \hat{y}_i|}{|y_i| + |\hat{y}_i|}, \quad (14)$$

### 4.3. Forecasting techniques included in comparison and hyperparameter optimization

Since KARN is essentially a recurrent extension of the basic KAN, similar to how RNNs were developed from MLPs, we included comparisons with three RNNs: Vanilla RNN, GRU, and LSTM to fairly assess its impact. The forecasting techniques for comparisons are given along with their architectural details in Table 2.

To ensure a fair comparison among all models, including our own, hyperparameter optimization was conducted using a grid search for each dataset and model. The hyperparameter search space is shown in Table 3.

For LSTM, GRU, and RNN models, we found that one recurrent layer was optimal, with most datasets converging at 64 units. However, the townhouse and office datasets converged at 128 units. Stochastic Gradient Descent (SGD) was the preferred optimizer in most cases, though for House 1, Adam proved to be the most effective. In our implementation of KARN, we found that a spline degree of 2 was optimal, and grid extension was applied only in the case of individual houses. Specifically, for House 3, the grid was extended to 10 points. For our model, we consistently used the MAE loss function across all cases, as MSE did not yield optimal performance. An early stopping mechanism was employed to monitor the validation loss, halting training after five consecutive epochs without improvement. The maximum number of epochs was set to 300, though training often concluded earlier when the early stopping criteria were met, with our proposed model typically converging before reaching 100 epochs. The learning rate was dynamically adjusted using the ReduceLROnPlateau schedule (Torghabeh et al., 2023), which reduces the rate when there is no observed improvement in performance. Weights were initialized using Xavier uniform initialization to optimize the training process (Kumar, 2017).

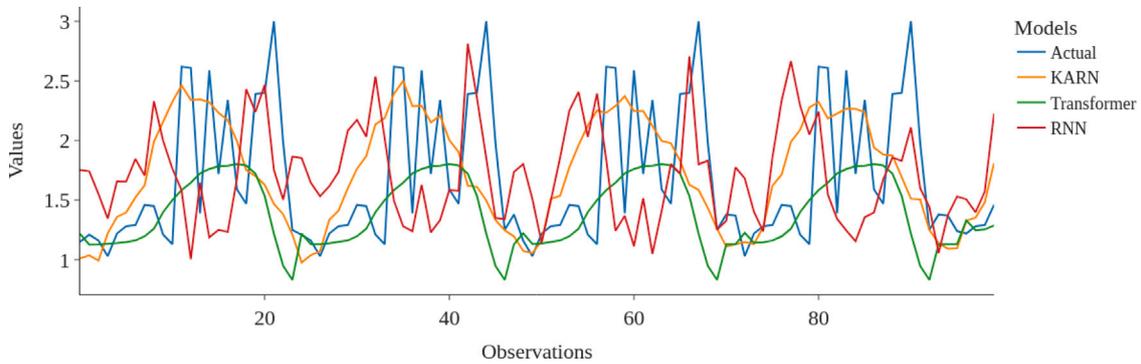

**Fig. 2.** House 2: Energy load characterized by observable random fluctuations, spikes and drops.

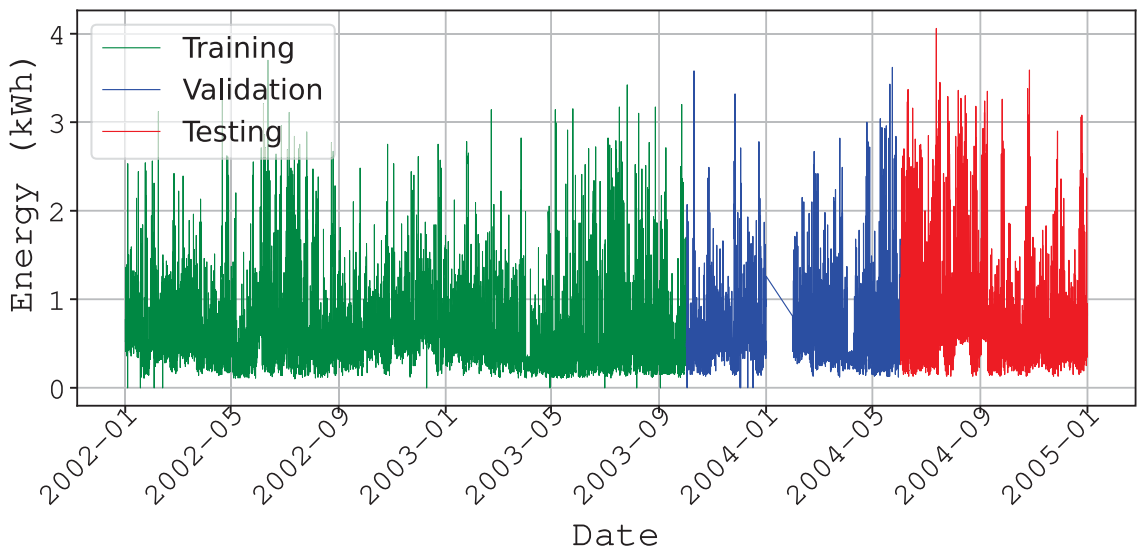

**Fig. 3.** House 1: Energy load characterized by sudden spikes, variations, and level shifts.

**Table 4**
Performance comparison for House 1 and House 2 (Testing days: House 1: 225 days, House 2: 43 days).

| Model | House 1 | | | House 2 | | |
|---|---|---|---|---|---|---|
| | MAE | RMSE | SMAPE | MAE | RMSE | SMAPE |
| RNN | 0.40 | 0.53 | 50.07% | 0.70 | 0.85 | 35.58% |
| GRU | 0.42 | 0.47 | 51.96% | 0.68 | 0.82 | 34.30% |
| LSTM | 0.39 | 0.49 | 48.28% | 0.63 | 0.82 | 32.53% |
| KARN | **0.36** | **0.45** | **44.88%** | **0.38** | **0.50** | **19.29%** |

## 5. Results and analysis

This section presents three case studies of building energy load forecasting: Case Study 1: Individual Houses; Case Study 2: Student Residences; and Case Study 3: Industrial and Commercial Buildings. These case studies ensure the deployment of models over diverse consumer types. The case studies will be followed by a discussion and an examination of the limitations.

### 5.1. Case study 1: Individual houses

This case study focuses on the energy consumption patterns observed in four individual houses, each presenting unique and diverse load profiles. These houses show a wide range of load behaviors, making forecasting more challenging. The diversity in consumption patterns can be seen in Figs. 3 to 6, where each house energy load is influenced by factors such as sudden spikes, variations, level shifts, and the presence of an electric vehicle in one case.

House 1 load patterns, as shown in Fig. 3, are highly irregular with spikes, drops, and random variations. This irregularity is due to the house belonging to an individual family whose routine varies. In this scenario, as given in Table 4, KARN outperformed all other models by achieving a SMAPE of 44.88%, compared to RNN's 50.07%, GRU's 51.96%, and LSTM's 48.28%. The results demonstrate that KARN can learn complex patterns better than other techniques due to its expanded learnable scale, which is a result of grid extension. These findings confirm that KARN is particularly well-suited to managing such complex datasets, outperforming even the more sophisticated architectures of LSTM and GRU.

For House 2, as seen in Fig. 4, the load is lower in April and May, while after June there is a level shift and large spike variations. In the test data, there are no trends or seasonality, only sudden spikes and drops. As depicted in Table 4, KARN outperformed all RNN variants. It achieved the lowest errors across all metrics, with a SMAPE of 19.29%, significantly lower than RNN's 35.58% and also better than both LSTM and GRU, which recorded SMAPE values of 32.53% and 34.30%, respectively. This once again suggests that KARN is particularly well-suited to handling datasets with clear patterns of drift and sudden changes. We presented actual versus predicted values in Fig. 2, which illustrates random fluctuations. It shows that, despite the concept drift in the training data, our proposed method maintained its performance compared to other models.

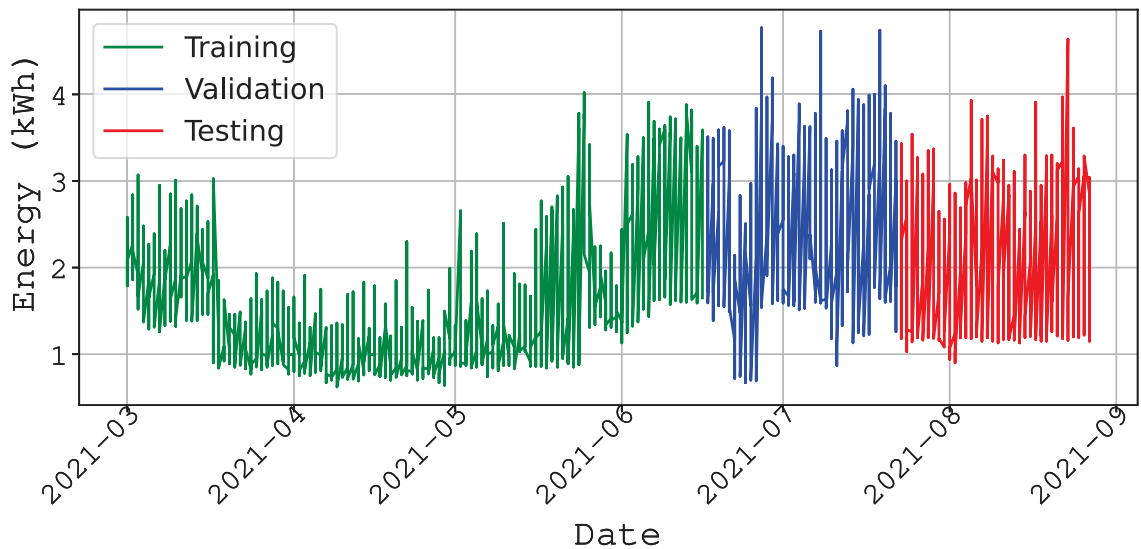

**Fig. 4.** House 2: Energy load characterized by concept drift, sudden spikes, cyclic variations, and level shifts.

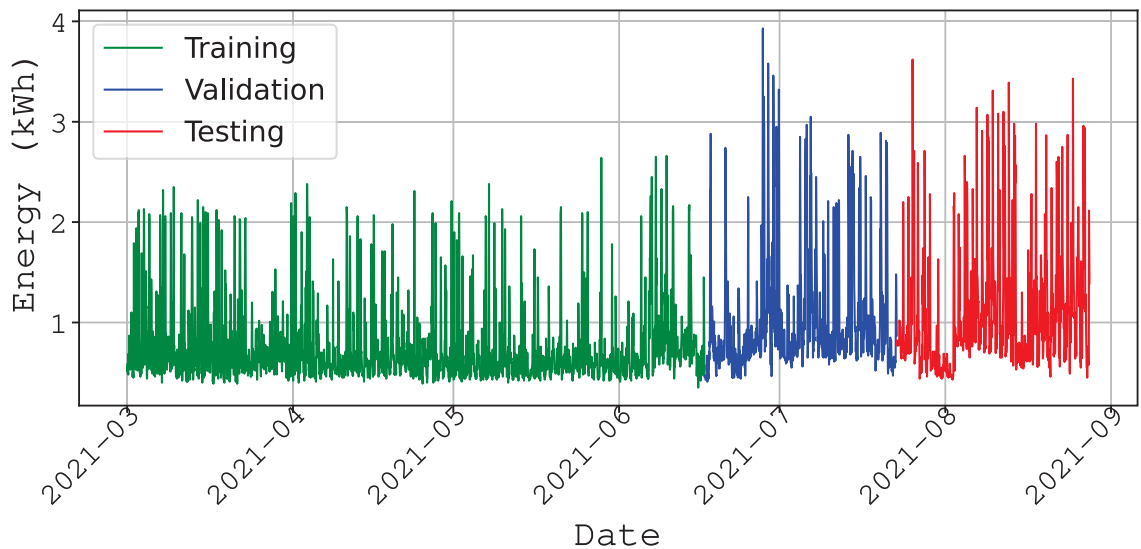

**Fig. 5.** House 3 (with eclectic vehicle): Energy load characterized by sudden spikes, cyclic variations, and level shifts.

**Table 5**
Performance Comparison for House 3 and House 4 (Testing days: House 3: 43 days, House 4: 43 days).

| Model | House 3 (EV) | | | House 4 | | |
|---|---|---|---|---|---|---|
| | MAE | RMSE | SMAPE | MAE | RMSE | SMAPE |
| RNN | 0.45 | 0.66 | 40.39% | 0.33 | 0.39 | 43.56% |
| GRU | 0.38 | **0.52** | 32.03% | 0.32 | 0.37 | 43.76% |
| LSTM | 0.37 | 0.53 | 31.71% | **0.28** | **0.36** | 39.75% |
| KARN | **0.35** | 0.54 | **28.73%** | **0.28** | 0.38 | **39.39%** |

In the case of House 3, we expanded our evaluation to include a house equipped with an Electric Vehicle (EV), which is expected to introduce additional complexity due to charging spikes. As illustrated in Fig. 5 and Table 5, these sudden spikes, along with level shifts, are clearly visible in the energy consumption pattern. Despite these challenges, our proposed KARN model outperformed both RNN and GRU, achieving a SMAPE of 28.73%, compared to RNN's 40.39%, LSTM's 31.71%, and GRU's 32.03%. These results highlight KARN's ability to manage complex, non-linear patterns in energy consumption data. However, it is important to note that GRU, with its gating mechanisms, delivered a slightly lower RMSE of 0.52, indicating that it remains a robust choice for this type of time series data. Overall, the performance of KARN in this scenario further validates its effectiveness in handling diverse and challenging energy consumption patterns, including those influenced by modern technologies like EVs.

House 4 further diversifies our consumer base by examining a townhouse rather than a standalone house. As seen in Fig. 6, sudden spikes are highly prevalent in the data. As seen from Table 5, once again, KARN demonstrated strong performance with a SMAPE of 39.39%, outperforming RNN 43.56%, LSTM 39.75%, and GRU 43.76%. However, LSTM narrowly outperformed KARN in terms of RMSE with an RMSE of 0.36, indicating that while KARN is effective, LSTM's architecture still holds an edge in certain scenarios.

The results, summarized in Tables 4 and 5, indicate that KARN consistently outperforms RNN across all houses, confirming its design is well suited for handling complex load data patterns. However, the advanced architectures of LSTM and GRU, in some cases, continue to show strong performance, particularly in datasets with more predictable patterns. Nonetheless, KARN's ability to handle more erratic and complex patterns with lower error rates than RNNs and close competition with LSTM and GRU highlights its potential as an effective forecasting model in diverse residential settings.

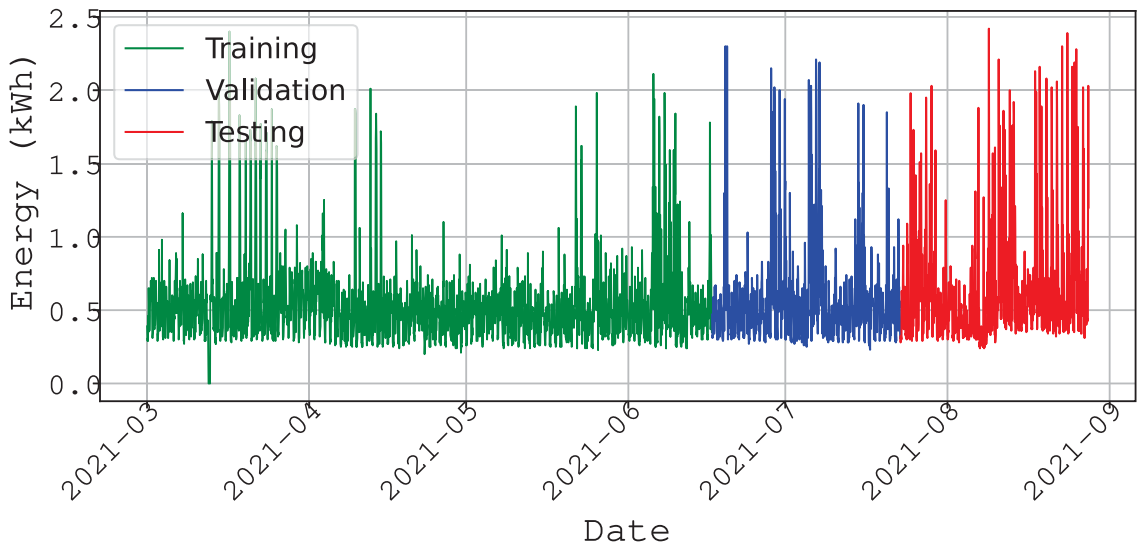

**Fig. 6.** House 4 (townhouse): Energy load characterized by sudden spikes, cyclic variations, and level shifts.

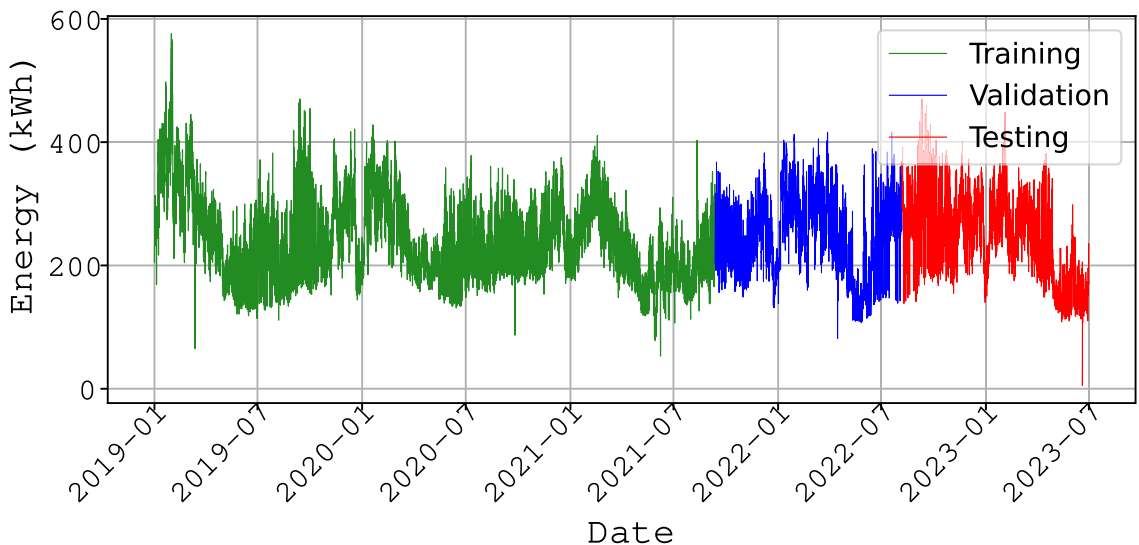

**Fig. 7.** Case Study 1: Residence 1 energy load patterns characterized by observable seasonal variations affected by students' routines.

### 5.2. Case study 2: Student residences

Case Study 1 focuses on two student residence buildings within a higher education institution. These residences are occupied by upper-year undergraduate students; thus, load patterns are influenced by students' academic routines. During summer months (May to August), patterns are likely to become more random due to the absence of regular occupants, as can be seen in Fig. 7. We can observe that energy usage is higher in some periods, such as early winter, and there are daily drops and spikes, indicating the influence of the university's daily routine.

As seen from Table 6, our proposed method, KARN, outperformed its equivalent variant in traditional ML technology, namely RNN,

**Table 6**
Performance comparison for Case study 1: Student residences (Testing: 335 days).

| Model | Residence 1 | | | Residence 2 | | |
|---|---|---|---|---|---|---|
| | MAE | RMSE | SMAPE | MAE | RMSE | SMAPE |
| RNN | 24.66 | 34.82 | 10.13% | 29.05 | 39.50 | 12.00% |
| GRU | 24.85 | 33.00 | 10.47% | **19.67** | 28.04 | **8.14%** |
| LSTM | **21.82** | 29.85 | **9.12%** | 21.76 | 30.33 | 9.29% |
| KARN | 22.63 | **29.70** | 9.40% | 20.57 | **28.04** | 8.93% |

achieving a SMAPE of 9.40% compared to RNN's 10.13%. This demonstrates that our design, which uses a layer structured similarly to RNN but with KAN non-linear weights, clearly outperforms RNN. However, LSTM outperformed our proposed KARN due to its effective gating structure, making it one of the standard algorithms for time series analysis. It is important to note that a fair comparison of KARN should be made with RNN rather than LSTM as KARN can be seen as RNN extension. Interestingly, KARN also outperformed GRU, which had a SMAPE of 10.47%. Overall, results from traditional RNNS and our proposed recurrent network show acceptable performance. We have included line plots in figures Fig. 8 and Fig. 10 to compare the actual and forecasted load values for KARN and other models, showing how accurately our model captures the patterns.

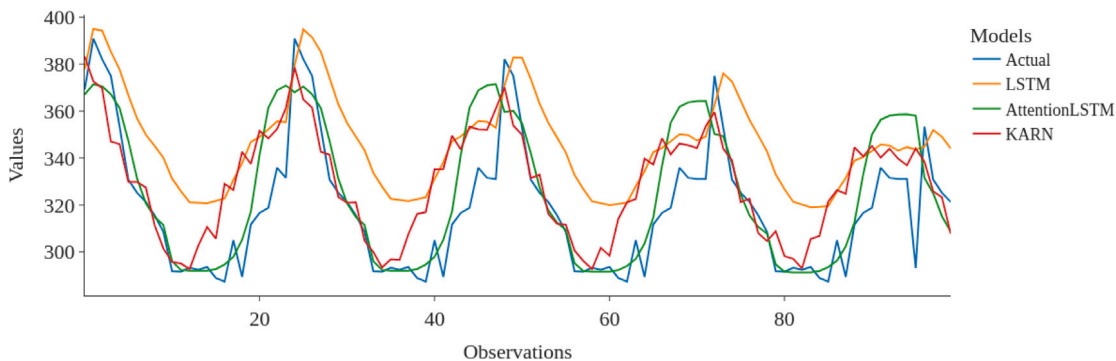

**Fig. 8.** Student Residence 1: Actual versus the forecasted load values for compared techniques.

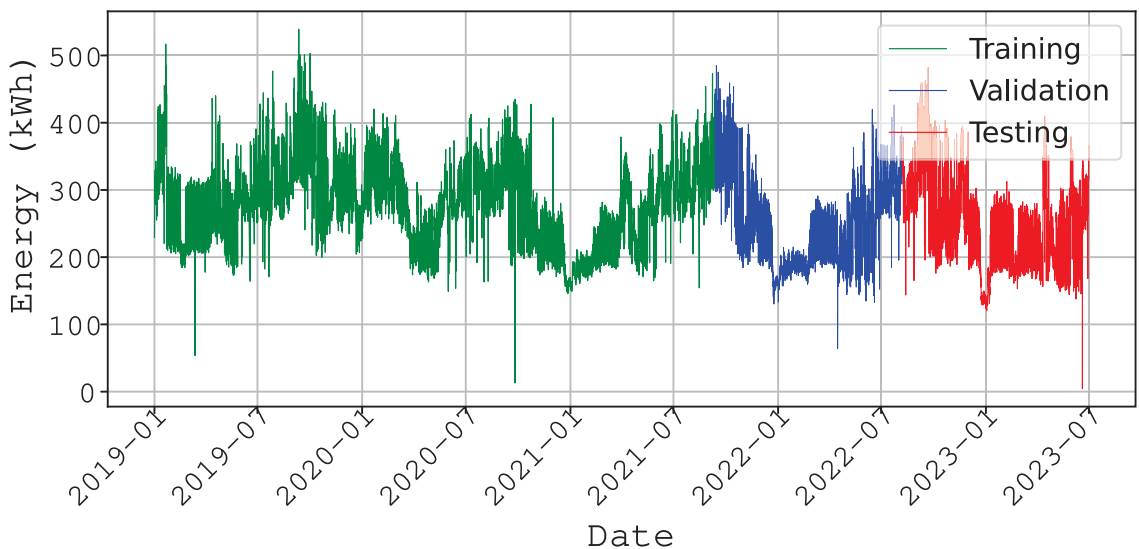

**Fig. 9.** Case Study 1: Residence 2 energy load characterized by observable seasonal variations affected by students' routines.

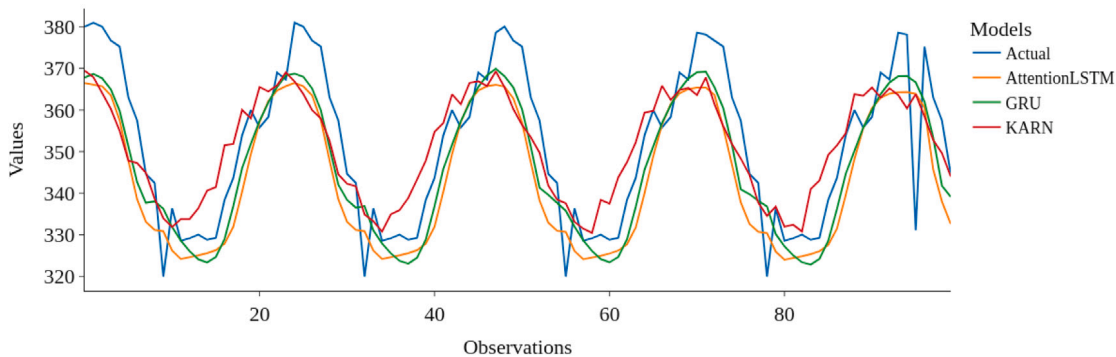

**Fig. 10.** Student Residence 2: Actual versus the forecasted load values for compared techniques.

In the case of Residence 2, we observe similar patterns, but they are more consistent, primarily due to students' routines. Once again, we can notice (Fig. 9) clear seasonality and consistent variations in the data that help models learn. There are some noticeable drops, likely due to student holidays and summer vacations. In this case, Table 6, KARN once again outperforms RNN by reducing the error to 8.93%, compared to RNN's 12.00%. However, possibly due to the more consistent nature of the data, GRU outperforms all other models, including ours and LSTM, by reducing the error to 8.14%. It is worth noting that the error in Residence 2 is lower than the errors in Residence 1. Importantly, in terms of the RMSE metric, our proposed KARN outperforms the others, indicating that large errors are clearly reduced by our model. Overall, the performance of all models is acceptable. We have also presented a line plot in Fig. 4 showing actual vs. forecasted load values for KARN and all other models, demonstrating how accurately our model captures the patterns.

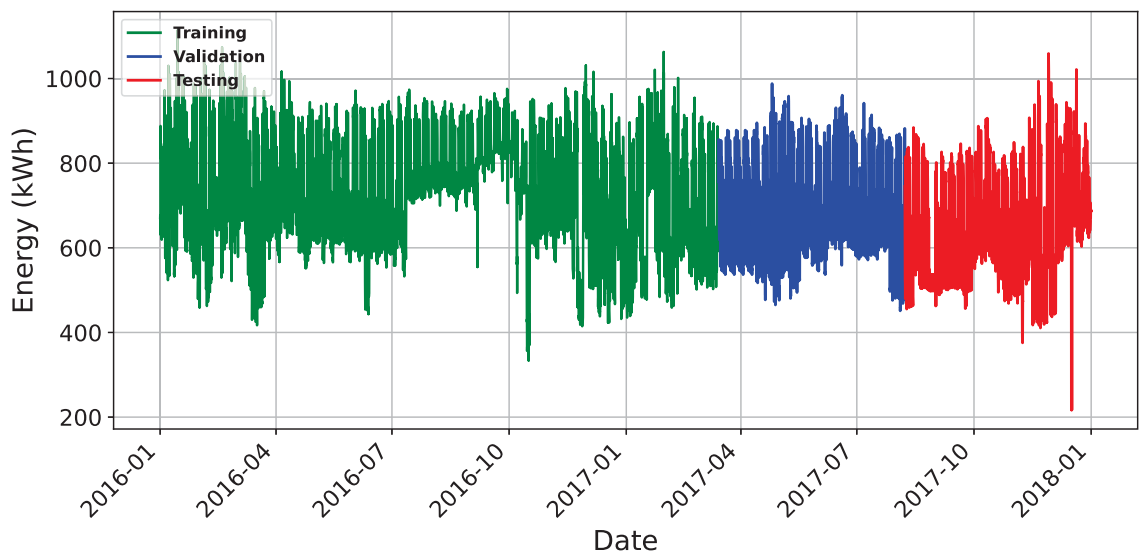

**Fig. 11.** Manufacturing: Energy load characterized by sudden drops, few spikes, variations, and level shifts.

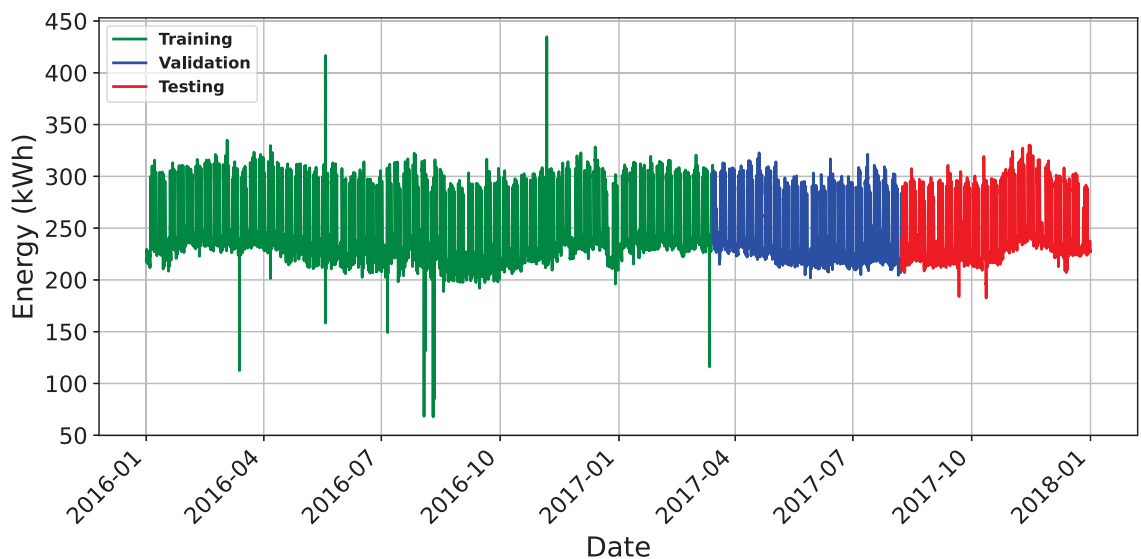

**Fig. 12.** Medical clinic: Energy load characterized by few drops and few spikes while consistent patterns.

Table 7
Performance comparison for manufacturing and medical clinic (Testing days: 146 days).

| Model | Manufacturing | | | Medical clinic | | |
|---|---|---|---|---|---|---|
| | MAE | RMSE | SMAPE | MAE | RMSE | SMAPE |
| RNN | 74.15 | 88.71 | 11.66% | 19.44 | 23.65 | 7.88% |
| GRU | 55.59 | 65.22 | 8.70% | 11.85 | 14.59 | 4.75% |
| LSTM | **48.45** | **56.87** | **7.67%** | 10.28 | 12.91 | 4.05% |
| KARN | 54.62 | 77.24 | 8.52% | **9.22** | **12.84** | **3.71%** |

### 5.3. Case study 3: Industrial and commercial buildings

This case study evaluates the energy consumption of four different types of industrial and commercial buildings: a manufacturing facility, a medical clinic, a retail store, and an office building. Each of these buildings presents unique load characteristics, as illustrated in Figs. 11 to 14. The performance of various forecasting models, including RNN, GRU, LSTM, and KARN, was analyzed across these settings, with results summarized in Table 7.

For the manufacturing building, as shown in Fig. 11, the energy load is characterized by sudden drops, occasional spikes, and level shifts, but an overall consistent pattern creates a favorable environment for accurate forecasting. As presented in Table 7, among the models evaluated, LSTM outperformed the others with the lowest MAE and SMAPE values of 48.45 and 7.67%, respectively. However, the KARN model also demonstrated competitive performance with a SMAPE of 8.52%, while RNN showed SMAPE of 11.66%. This indicates that KARN again outperforms its traditional analog, RNN, highlighting its robustness in handling complex patterns typical of industrial settings.

In the case of the medical clinic, the energy consumption pattern, depicted in Fig. 12, is relatively stable with few spikes or drops. As shown in Table 7, this consistency in load patterns is reflected in the model performance. KARN achieved the best results across all metrics, with a notably low SMAPE of 3.71%, compared to LSTM's 4.05%, GRU's 4.75%, and RNN's 7.88%. This demonstrates KARN's ability to effectively model the more predictable energy usage typical of medical facilities, outperforming its neural network counterparts.

The retail store, shown in Fig. 13, presents a more volatile load profile, with sudden drops, although most patterns remain consistent. As presented in Table 8, KARN again demonstrated superior performance, achieving the lowest SMAPE of 7.17%, which was slightly better than LSTM's 7.56%, GRU's 7.57%, and RNN's 10.10%. However, in terms of RMSE, LSTM slightly outperformed KARN by achieving an RMSE of 26.67. This suggests that while KARN is well-suited to capturing the intricate patterns in retail energy consumption, LSTM still holds an important advantage in certain aspects of prediction accuracy.

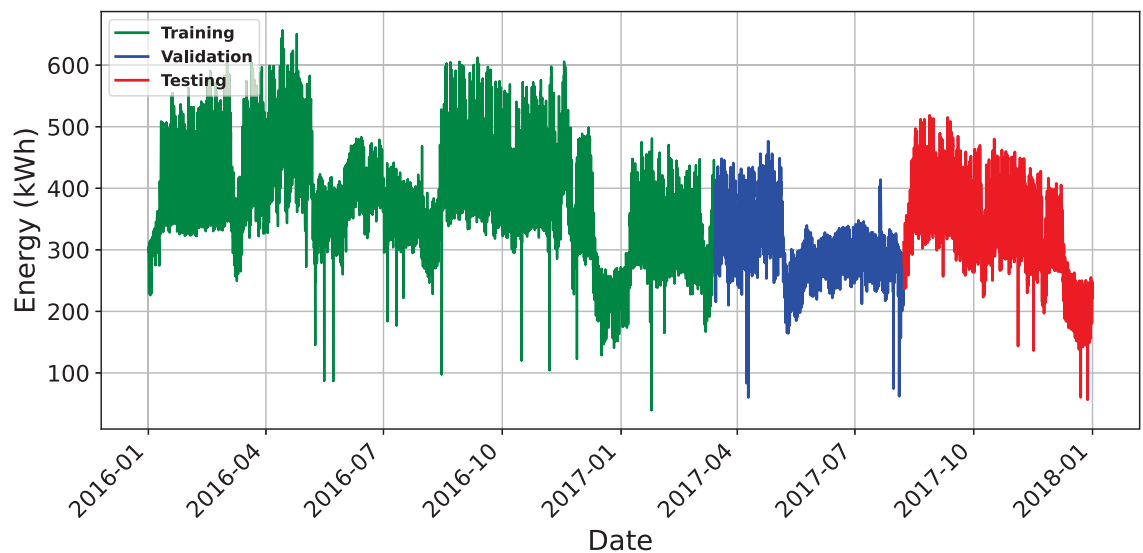

**Fig. 13.** Retail store: Energy load characterized by sudden spikes and drops and cyclic variations.

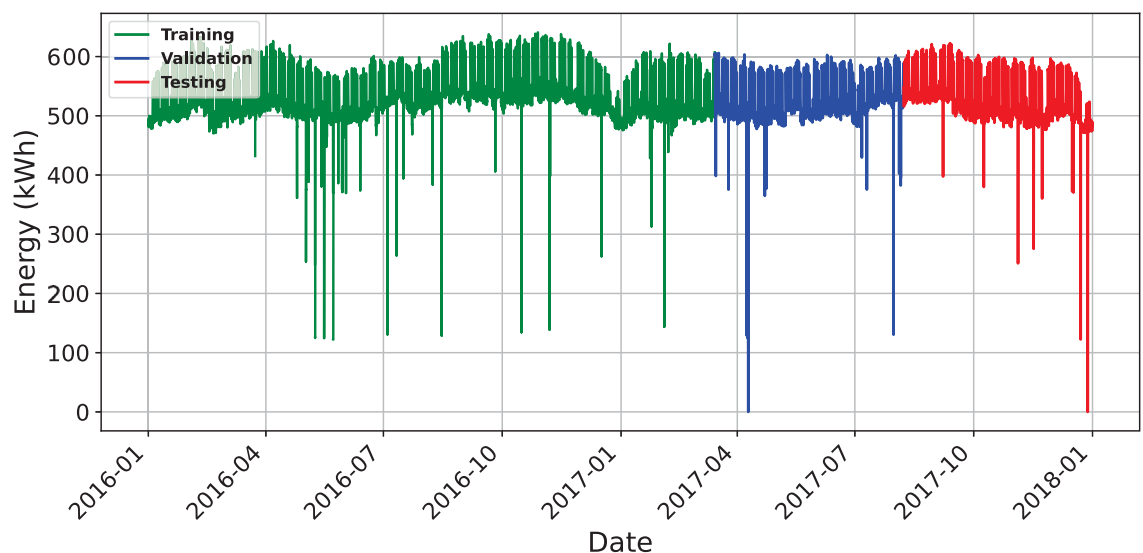

**Fig. 14.** Office: Energy load characterized by sudden drops, cyclic variations, and level shifts.

The office building, characterized by sudden drops and cyclic variations as seen in Fig. 14, presented a more stable environment compared to the other buildings. As presented in Table 8, KARN outperformed all other models with the best results across all metrics, including a SMAPE of 3.55%, compared to LSTM's 3.77%, GRU's 3.98%, and RNN's 4.18%. This consistent performance across various metrics underscores KARN's strength in environments with more regular energy usage patterns, such as those typically seen in office settings.

In summary, across all four types of buildings, KARN consistently delivered strong performance in environments with both stable and complex energy usage patterns. Its competitive results in more volatile settings, such as the retail store, also highlight its versatility. These findings suggest that KARN is a highly effective model for energy load forecasting in diverse industrial and commercial environments, often outperforming more established models like LSTM and GRU. KARN demonstrated superior performance in most scenarios, achieving the lowest SMAPE values for the office building 3.55%, retail store 7.17%, and manufacturing building 8.52%. While LSTM slightly outperformed KARN in terms of RMSE for the retail store, KARN's overall consistency across various metrics and building types emphasizes its robustness and adaptability. This comprehensive evaluation across different building types and energy consumption patterns indicates that KARN is a promising approach for short-term load forecasting in various commercial and industrial settings. Its ability to capture both regular patterns and sudden variations makes it a valuable tool for energy management and planning in diverse environments.

### 5.4. Discussion

It is also important to evaluate the computational efficiency of the KARN to examine its applicability in such diverse settings. Tables 9 and 10 provides a comparison of training and testing times across different models, including KARN, LSTM, GRU, and RNN. These results were obtained using a computer system with an AMD Ryzen Threadripper PRO 5955WX processor and an NVIDIA GA102GL RTX A6000 GPU. While KARN shows higher computational complexity, denoted by its $O(Gn^2)$ complexity, compared to traditional recurrent models, it maintains competitive inference times, with a maximum testing duration of 0.18 min for the largest dataset. Although KARN's computational demands are higher than those of LSTM, this increase is still acceptable and feasible for deployment in energy systems.

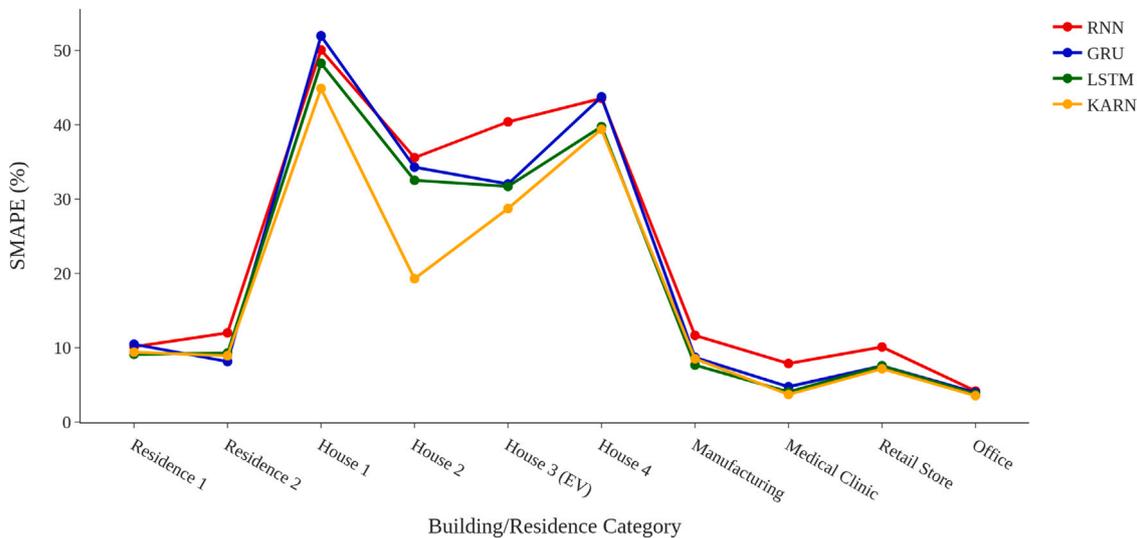

**Fig. 15.** Comparison of KARN with three approaches for each student residence, individual houses, industrial and commercial consumers. Note that the best approaches are not the same across the consumers, but for all consumers, KARN achieves the lowest SMAPE.

**Table 8**
Performance comparison for retail store and office (Testing days: 146 days).

| Model | Retail Store | | | Office | | |
|---|---|---|---|---|---|---|
| | MAE | RMSE | SMAPE | MAE | RMSE | SMAPE |
| RNN | 33.36 | 38.57 | 10.10% | 20.90 | 24.64 | 4.18% |
| GRU | 23.96 | 28.67 | 7.57% | 20.07 | 23.11 | 3.98% |
| LSTM | 22.25 | **26.67** | 7.56% | 19.13 | 22.54 | 3.77% |
| KARN | **21.96** | 28.83 | **7.17%** | **17.69** | **21.47** | **3.55%** |

**Table 9**
Comparison of computation (in min) and complexity for residences and houses.

| Model | Complexity | Residence 1 | | House 1 | |
|---|---|---|---|---|---|
| | | Training | Testing | Training | Testing |
| RNN | $O(2n)$ | 20.0 | 0.10 | 14.5 | 0.06 |
| LSTM | $O(4n)$ | 39.9 | 0.13 | 26.2 | 0.08 |
| GRU | $O(3n)$ | 30.0 | 0.12 | 20.0 | 0.06 |
| KARN | $O(Gn^2)$ | 70.5 | 0.18 | 60.3 | 0.10 |

**Table 10**
Comparison of computation (in min) and complexity for House 2 and manufacturing.

| Model | Complexity | House 2 | | Manufacturing | |
|---|---|---|---|---|---|
| | | Training | Testing | Training | Testing |
| RNN | $O(2n)$ | 6.1 | 0.02 | 10.0 | 0.03 |
| LSTM | $O(4n)$ | 8.6 | 0.02 | 15.2 | 0.04 |
| GRU | $O(3n)$ | 7.1 | 0.02 | 12.0 | 0.03 |
| KARN | $O(Gn^2)$ | 18.5 | 0.03 | 40.1 | 0.08 |

KARN's sophistication arises from its use of temporal spline functions and learnable activation mechanisms on edges rather than nodes, which allows for more refined modeling of non-linear dependencies in the data. While this increased complexity results in longer training times, it contributes to KARN's superior forecasting accuracy. In six out of ten consumer types, KARN outperforms LSTM, GRU, and RNN, and it surpasses RNN in all ten buildings, as evidenced by the results presented in Tables 6, 4, 5, 8 and 7. Fig. 15 compare the proposed methods with the top three other approaches. This versatility is important for real-world applications, where energy consumption patterns can vary across consumer types, from student residences to industrial facilities. As shown in the case studies, KARN consistently outperformed traditional models, especially in environments characterized by irregular load patterns, such as individual houses and industrial buildings. However, in some cases, such as residences and manufacturing datasets, LSTM or GRU outperformed KARN, which can be attributed to their well-established memory systems based on gating mechanisms. The grid extension mechanism employed by KARN proved to be particularly effective in these settings, enabling the model to capture sudden spikes and level shifts more accurately than other models.

KARN is sensitive to its hyperparameter settings, such as spline degree and grid size, which require careful tuning. In contrast, models like LSTM and GRU are generally more robust to variations in hyperparameters. Additionally, KARN mitigates issues such as vanishing gradients, exploding gradients, and overfitting due to its design, particularly by placing activations on weights. KARN, in simple terms, is an RNN built using the Kolmogorov–Arnold representation theorem. Despite its complexity, as shown in the results, KARN delivers performance better or comparable to that of architectures like LSTM and GRU, which typically require multiple recurrent layers to achieve similar results. In summary, KARN represents a significant advancement in load forecasting across diverse consumer types. While the model's increased complexity and computational demands are noteworthy, its ability to consistently outperform traditional models in complex scenarios highlights its potential as a powerful tool for energy management and planning.

## 6. Conclusion

This paper introduced KARN, a novel load forecasting approach designed to handle the complex and varied energy consumption patterns of diverse consumer types. KARN leverages the strengths of KANs, redesigned to capture the memory of previous long-time steps and to enhance the model's ability to capture temporal dependencies in energy data. It utilizes learnable spline functions and non-linear weights, which offer a more flexible and accurate way to model non-linear relationships in load data.

Learnable spline functions and edge-based activations in KARN reduce the possibility of vanishing and exploding gradients. KARN offers a more flexible and accurate approach to modeling non-linear relationships in load data, setting it apart from traditional recurrent models like RNN, LSTM, and GRU. The effectiveness of KARN was rigorously evaluated across ten diverse real-world datasets such as student residences, individual houses, and various industrial and commercial buildings. The results consistently demonstrated that KARN outperformed its traditional counterpart RNN in all cases, and also outperformed LSTM and GRU in six buildings. Notably, KARN achieved lower SMAPE values across most datasets, highlighting its robustness and adaptability in forecasting tasks.

This study also set out to investigate the research question: *Can a single forecasting model effectively adapt to and capture the diverse energy consumption patterns demonstrated across various consumer groups?* The findings indicate that the KARN addresses this by consistently outperforming its traditional counterpart, Vanilla RNN, across all ten datasets. The KARN also outperformed LSTM and GRU in six out of ten datasets. For instance, in the case of House 2 and the Retail Store, KARN achieved approximately 11% lower SMAPE values than LSTM, which shows its spline-based learning ability to handle sudden spikes and level shifts. As described in case studies 1 and 3, KARN performed well in environments with irregular load patterns, such as individual houses and industrial buildings. LSTM and GRU achieved better results in more stable datasets, such as student residences and manufacturing datasets. In contrast, KARN shows a more consistent performance across diverse datasets, which makes it a reliable choice for complex energy consumption scenarios.

While KARN's advanced architecture and grid extension mechanism contribute to its superior performance, they also introduce increased computational complexity, leading to longer training times. Despite this, KARN's inference time remained competitive, making it a viable option for practical deployment in real-world scenarios. KARN represents a significant advancement in the field of load forecasting, offering a powerful tool for managing energy consumption across diverse consumer types. Its ability to effectively model both stable and volatile energy patterns underscores its potential for widespread application in energy management and planning.

The study also identified potential areas for improvement, including the need for careful hyperparameter tuning and the development of more efficient training algorithms to mitigate the model's computational demands. Future work will also investigate the potential of transfer learning to enhance model generalization across similar consumer types.

## CRediT authorship contribution statement

**Muhammad Umair Danish:** Writing – original draft, Visualization, Validation, Software, Methodology, Conceptualization. **Katarina Grolinger:** Writing – review & editing, Validation, Investigation, Funding acquisition, Data curation.

## Declaration of competing interest

The authors declare the following financial interests/personal relationships which may be considered as potential competing interests: Katarina Grolinger reports financial support was provided by Natural Sciences and Engineering Research Council of Canada. Katarina Grolinger reports financial support was provided by Environment and Climate Change Canada. If there are other authors, they declare that they have no known competing financial interests or personal relationships that could have appeared to influence the work reported in this paper.

## Data availability

The data that has been used is confidential.

## References

Aghaei, A.A., 2024. rKAN: Rational Kolmogorov–Arnold networks. arXiv preprint arXiv:2406.14495.

Ahmed, Z., Jamil, M., 2024. Campus electric load forecasting using recurrent neural networks. In: 2024 12th International Conference on Smart Grid. IcSmartGrid, IEEE.

Azam, B., Akhtar, N., 2024. Suitability of KANs for computer vision: A preliminary investigation. arXiv preprint arXiv:2406.09087.

Bessani, M., Massignan, J.A., Santos, T.M., London, Jr., J.B., Maciel, C.D., 2020. Multiple households very short-term load forecasting using Bayesian networks. Electr. Power Syst. Res..

Business Standard News, 2024. World news electricity demand. URL https://shorturl.at/Tx5BG.

Cai, M., Pipattanasomporn, M., Rahman, S., 2019. Day-ahead building-level load forecasts using deep learning vs. traditional time-series techniques. Appl. Energy.

Chen, T., Chen, G., Chen, W., Hou, S., Zheng, Y., He, H., 2021. Application of decoupled ARMA model to modal identification of linear time-varying system based on the ICA and assumption of "short-time linearly varying". J. Sound Vib..

Cheon, M., 2024. Kolmogorov–Arnold network for satellite image classification in remote sensing. arXiv preprint arXiv:2406.00600.

Danish, M.U., Grolinger, K., 2024. Leveraging hypernetworks and learnable kernels for consumer energy forecasting across diverse consumer types. IEEE Trans. Power Deliv..

De Boor, C., 1972. On calculating with B-splines. J. Approx. Theory.

Deng, S., Chen, F., Dong, X., Gao, G., Wu, X., 2021. Short-term load forecasting by using improved GEP and abnormal load recognition. ACM Trans. Internet Technol. (TOIT).

European Climate, Energy and Environment, 2024. 2030 climate & energy framework. URL https://climate.ec.europa.eu/eu-action/climate-strategies-targets/2030-climate-energy-framework_en. (Accessed 14 August 2022).

Fekri, M., Grolinger, K., Mir, S., 2023. Asynchronous adaptive federated learning for distributed load forecasting with smart meter data. Int. J. Electr. Power Energy Syst. accepted.

Genet, R., Inzirillo, H., 2024a. A temporal Kolmogorov–Arnold transformer for time series forecasting. arXiv preprint arXiv:2406.02486.

Genet, R., Inzirillo, H., 2024b. Tkan: Temporal Kolmogorov–Arnold networks. arXiv preprint arXiv:2405.07344.

Giacomazzi, E., Haag, F., Hopf, K., 2023. Short-term electricity load forecasting using the temporal fusion transformer: Effect of grid hierarchies and data sources. In: Proceedings of the 14th ACM International Conference on Future Energy Systems.

Gong, H., Rallabandi, V., McIntyre, M.L., Hossain, E., Ionel, D.M., 2021. Peak reduction and long term load forecasting for large residential communities including smart homes with energy storage. IEEE Access.

He, X., Zhao, W., Gao, Z., Zhang, Q., Wang, W., 2024. A hybrid prediction interval model for short-term electric load forecast using holt-winters and gate recurrent unit. Sustain. Energy Grids Netw..

Kiamari, M., Kiamari, M., Krishnamachari, B., 2024. GKAN: Graph Kolmogorov–Arnold networks. arXiv preprint arXiv:2406.06470.

Kong, W., Dong, Z.Y., Jia, Y., Hill, D.J., Xu, Y., Zhang, Y., 2017. Short-term residential load forecasting based on LSTM recurrent neural network. IEEE Trans. Smart Grid.

Kumar, S.K., 2017. On weight initialization in deep neural networks. arXiv preprint arXiv:1704.08863.

L'Heureux, A., Grolinger, K., Capretz, M., 2022. Transformer-based model for electrical load forecasting. Energies.

Li, S., Kong, X., Yue, L., Liu, C., Khan, M.A., Yang, Z., Zhang, H., 2023a. Short-term electrical load forecasting using hybrid model of manta ray foraging optimization and support vector regression. J. Clean. Prod..

Li, Z., Liu, F., Yang, W., Peng, S., Zhou, J., 2021. A survey of convolutional neural networks: analysis, applications, and prospects. IEEE Trans. Neural Netw. Learn. Syst..

Li, Y., Zhu, N., Hou, Y., 2023b. A novel hybrid model for building heat load forecasting based on multivariate empirical modal decomposition. Build. Environ..

Liu, Z., Wang, Y., Vaidya, S., Ruehle, F., Halverson, J., Soljačić, M., Hou, T.Y., Tegmark, M., 2024. Kan: Kolmogorov–Arnold networks. arXiv preprint arXiv:2404.19756.

Mao, W., Yu, S., Chen, W., 2024. Short-term power load forecasting method based on GRU-transformer combined neural network model. J. Comput. Inf. Technol..

Muzaffar, S., Afshari, A., 2019. Short-term load forecasts using LSTM networks. Energy Procedia 158, 2922–2927.

Qin, J., Zhang, Y., Fan, S., Hu, X., Huang, Y., Lu, Z., Liu, Y., 2022. Multi-task short-term reactive and active load forecasting method based on attention-LSTM model. Int. J. Electr. Power Energy Syst..

Rezaei, E., Dagdougui, H., 2020. Optimal real-time energy management in apartment building integrating microgrid with multizone HVAC control. IEEE Trans. Ind. Inform..

Skala, R., Elgalhud, M.A.T., Grolinger, K., Mir, S., 2023. Interval load forecasting for individual households in the presence of electric vehicle charging. Energies.

Tan, S., Yang, Y., Zhang, Y., 2024. Short-term power load forecasting using informer encoder and bi-directional LSTM. In: E3S Web of Conferences. EDP Sciences.

Tang, T., Chen, Y., Shu, H., 2024. 3D U-KAN implementation for multi-modal MRI brain tumor segmentation. arXiv preprint arXiv:2408.00273.
Torghabeh, F.A., Modaresnia, Y., Khalilzadeh, M.M., 2023. Effectiveness of learning rate in dementia severity prediction using VGG16. Biomed. Eng.: Appl. Basis Commun..
Triebe, O., Laptev, N., Rajagopal, R., 2019. AR-Net: A simple auto-regressive neural network for time-series. arXiv preprint arXiv:1911.12436.
Vaca-Rubio, C.J., Blanco, L., Pereira, R., Caus, M., 2024. Kolmogorov-Arnold networks (kans) for time series analysis. arXiv preprint arXiv:2405.08790.
Vaswani, A., Shazeer, N., Parmar, N., Uszkoreit, J., Jones, L., Gomez, A.N., Kaiser, Ł., Polosukhin, I., 2017. Attention is all you need. Adv. Neural Inf. Process. Syst. 30.
Wei, N., Yin, C., Yin, L., Tan, J., Liu, J., Wang, S., Qiao, W., Zeng, F., 2024. Short-term load forecasting based on WM algorithm and transfer learning model. Appl. Energy.
Wu, W., Peng, M., 2017. A data mining approach combining $K$-means clustering with bagging neural network for short-term wind power forecasting. IEEE Internet Things J..
Xu, K., Chen, L., Wang, S., 2024. Kolmogorov–Arnold networks for time series: Bridging predictive power and interpretability. arXiv preprint arXiv:2406.02496.
Yamak, P.T., Yujian, L., Gadosey, P.K., 2019. A comparison between ARIMA, LSTM, and GRU for time series forecasting. In: 2nd International Conference on Algorithms, Computing and Artificial Intelligence.
Yao, X., Fu, X., Zong, C., 2022. Short-term load forecasting method based on feature preference strategy and LightGBM-XGboost. IEEE Access.
Yu, Y., Si, X., Hu, C., Zhang, J., 2019. A review of recurrent neural networks: LSTM cells and network architectures. Neural Comput..
Zeng, A., Chen, M., Zhang, L., Xu, Q., 2023. Are transformers effective for time series forecasting? In: Proceedings of the AAAI Conference on Artificial Intelligence.
Zhang, X., Chan, K.W., Li, H., Wang, H., Qiu, J., Wang, G., 2020. Deep-learning-based probabilistic forecasting of electric vehicle charging load with a novel queuing model. IEEE Trans. Cybern..
Zhang, Y.-F., Chiang, H.-D., 2019. Enhanced ELITE-load: A novel CMPSOATT methodology constructing short-term load forecasting model for industrial applications. IEEE Trans. Ind. Inform..
Zhang, G., Wei, C., Jing, C., Wang, Y., 2022. Short-term electrical load forecasting based on time augmented transformer. Int. J. Comput. Intell. Syst..